\documentclass{llncs}
\usepackage{makeidx}  % allows for indexgeneration
\usepackage{graphicx}
\usepackage{tabularx}
\usepackage{lscape}
\usepackage{amsmath}
\usepackage{bm}
\usepackage[table,xcdraw]{xcolor}
\usepackage{url}
\usepackage{subcaption}
\usepackage{tabularx}
\usepackage{caption}
\usepackage{multirow}
\usepackage{etoolbox} % for "\patchcmd" macro
\makeatletter
\patchcmd{\ps@headings}{\rlap{\thepage}}{}{}{}
\patchcmd{\ps@headings}{\llap{\thepage}}{}{}{}
\makeatother

\begin{document}
%
%\frontmatter          % for the preliminaries
%
%\pagestyle{headings}  % switches on printing of running heads
%\addtocmark{Hamiltonian Mechanics} % additional mark in the TOC

%\tableofcontents
%
\mainmatter              % start of the contributions
\title{Digital Twins for Human-Robot Collaboration: A Future Perspective}
\titlerunning{Digital Twins for Human-Robot Collaboration}  % abbreviated title (for running head)
\author{Mohamad Shaaban, Alessandro~Carf\`i, \and Fulvio~Mastrogiovanni}
\authorrunning{Mohamad Shaaban et al.} % abbreviated author list (for running head)
\tocauthor{Mohamad Shaaban, Alessandro~Carf\`i, and Fulvio~Mastrogiovanni}
\institute{Department of Informatics, Bioengineering, Robotics, and Systems Engineering,
University of Genoa, Via All'opera Pia 13, 16145 Genova, Italy\\
\email{mohamad.shaaban@edu.unige.it}\\}

© 2023 Springer. Personal use of this material is permitted. Permission from Springer must be obtained for all other uses, in any current or future media, including reprinting/republishing this material for advertising or promotional purposes, creating new collective works, for resale or redistribution to servers or lists, or reuse of any copyrighted component of this work in other works.
\newpage
\maketitle              % typeset the title of the contribution

\begin{abstract}
As collaborative robot (Cobot) adoption in many sectors grows, so does the interest in integrating digital twins in human-robot collaboration (HRC). Virtual representations of physical systems (PT) and assets, known as digital twins, can revolutionize human-robot collaboration by enabling real-time simulation, monitoring, and control. In this article, we present a review of the state-of-the-art and our perspective on the future of digital twins (DT) in human-robot collaboration. We argue that DT will be crucial in increasing the efficiency and effectiveness of these systems by presenting compelling evidence and a concise vision of the future of DT in human-robot collaboration, as well as insights into the possible advantages and challenges associated with their integration.
\keywords{Digital Twin, Human-Robot Collaboration, Cyber-Physical Systems }
\end{abstract}

\section{Introduction}
 \begin{figure*}[t]
\centering 
\includegraphics[width=0.95 \textwidth]{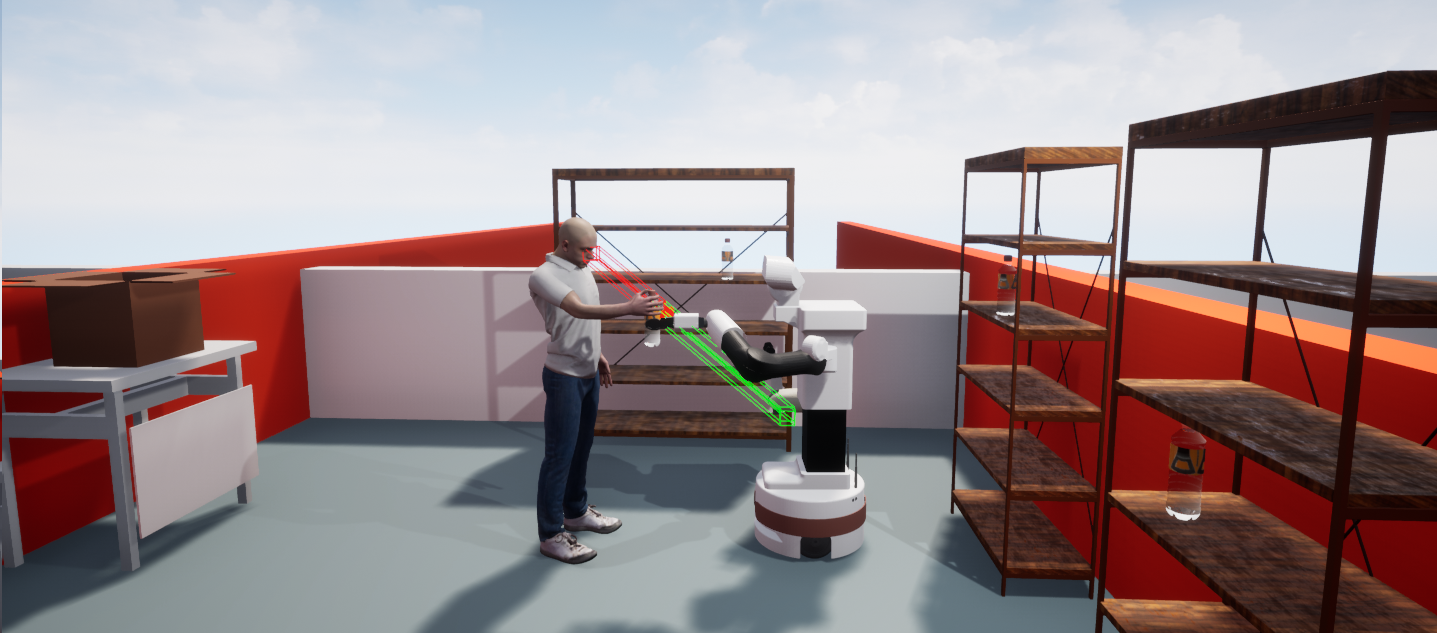}
\caption{The virtual representation of a human-robot interaction inside a DT.}
\label{img:dt-types}
\end{figure*}
%
%As the industry evolves toward personalized products and services, the product lifetime is growing shorter \cite{ourtzis:d}. However, as profit margins shrink and established processes struggle to meet the demand for personalized  items, this development poses issues for businesses \cite{Chryssolouris:g:ms}. While robots provide speed and efficiency, they cannot replace human characteristics such as cognitive capacity and adaptability, on the other hand, human labor is expensive. Collaborative robots, which combine the advantages of human skills and robots' speed, may give a solution. Cobots are utilized in HRC to accomplish the appropriate amount of collaboration and safety, and here is where DT comes into play. A digital twin (DT) is a virtual representation of the work environment, that can be used online or offline to simulate the real-world characteristics and behavior of a product, process, or service \cite{Bilberg:A:DT}. DTs are gaining popularity as a data augmentation and representation medium. Advances in computing power and connection are driving the expansion of DT in HRC.

 %They enable real-time monitoring, analysis, and design of human-robot  interactions, assisting enterprises in optimizing processes and increasing  overall efficiency. The use of DTs in HRC has the potential to transform how businesses approach their production processes, providing new opportunities to achieve better results while lowering costs and increasing the customer experience.

As the industry evolves toward customization of products and services, the product lifetime is shortening \cite{ourtzis:d}. Furthermore, as profit margins shrink, established processes struggle to meet the demand for product personalization, generating new challenges for industries \cite{Chryssolouris:g:ms}. In this context, robotization is a great opportunity, but while robots provide speed and efficiency, they
 cannot match human cognitive capabilities and flexibility. The research on collaborative robotics generates from these considerations and forecasts a close interaction of humans and robots to maintain high production standards while increasing flexibility in the production line. Nevertheless, the coexistence of humans and robots in the same environment poses challenges to safety and efficiency. Research in this domain is vast and, over time, addressed different subproblems characterizing human-robot collaboration (HRC) (e.g., efficient communication \cite{maccio2022mixed}, flexible task-motion planning \cite{darvish2020hierarchical}, and physical safety \cite{Kousi::Niki}). Recently researchers started investigating the possible application of digital twins to the HRC \cite{Malik::Brem}. A digital twin (DT) is a virtual representation of the work environment simulating the real-world characteristics and behavior of a product, process, or service \cite{Bilberg:A:DT}. The rise of DT is driven by improvements in simulation technologies, an overall increase in available computation power, and the perceived potential of this tool. Digital twins, in fact, would enable real-time monitoring, analysis, and design of human-robot interactions unlocking new application potential to increase productivity while promoting the well-being of the workers. The state-of-the-art on DT applications in HRC is still new and unstructured. With this article, we aim to provide a formalization of the main concepts related to DT, a brief review of the literature, and our own perspective on the opportunities and challenges of DT in HRC.

The rest of the article is structured as follows. In Section \ref{def}, we provide definitions of the main concepts related to DT in HRC. In Section \ref{taxonomy}, we formalize a taxonomy for digital twins in HRC scenarios. In Section \ref{review}, we extensively analyze the current literature. Finally, we present our discussion and conclusions.
\section{Definitions}
\label{def}

Before delving into the literature analysis on the usage of digital twins in human-robot collaboration, we need to define some terminology and concepts. We will do so by examining the state-of-the-art on digital twins more generally and discussing how these concepts could relate to the HRC scenario.
%\textcolor{red}{In order to establish a solid foundation, the key concepts related to this field will be defined first. The section will then delve into a comprehensive overview of digital twins, including the necessary technical components and the level of data integration involved. Subsequently, the criteria used for search and selection will be presented, followed by its outcomes.}

\subsection{Digital Twin}
A Digital Twin (DT) is a digital representation of a physical system. Michael Grieves and John Vickers, in 2003, first introduced the DT concept as the foundation for product life-cycle management \cite{Kritzinger:Werner}. Later, Grieves formalized the DT definition by identifying its foundational components: the physical entity, the virtual representation, and the stream of information that ties the two \cite{Grieves::DT::M}. The first component groups all the agents and tools from the real world that are of interest for the virtual representation. In HRC, the physical entity may include human operators, robots, and their shared working environment. The virtual representation creates a version of the physical system thanks to mathematical models. A virtual representation is not only a way to display the information collected from the real world but can also complement it with its internal model, for example, to build inferential modules (e.g., fatigue estimation \cite{merlo2023ergonomic} and activity recognition \cite{carfi2021gesture}) or decision strategies (e.g., task or motion planning \cite{darvish2020hierarchical}).

\begin{figure*}[t]
\centering 
\includegraphics[width=0.9\textwidth]{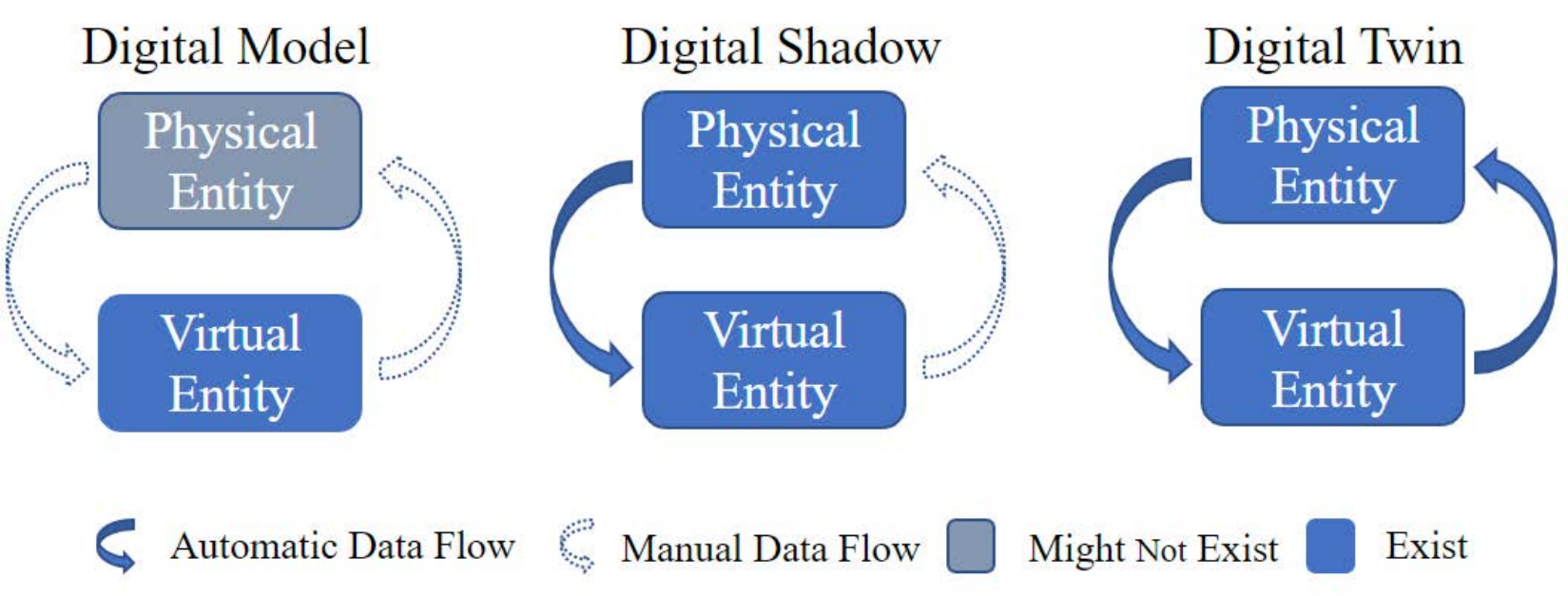}
\caption{Digital twin classification according to the level of data integration.}
\label{img:dt-types1}
\end{figure*}

\subsection{Level of Data Integration}
Michael Grives \cite{Grieves::DT::M}, in his digital twin formalization, ascribed a fundamental role to the communication between the physical entity and the virtual representation. This communication layer describes how the data flows from the real to the virtual world and vice-versa. Despite the DT definition imposing a bidirectional communication between the physical entity and the virtual representation, often in the literature, intermediated solutions are adopted. Therefore, many researchers identified three categories differing in the level of data integration between the digital replica and the physical entity: digital twins (DT), digital models (DM), and digital shadows (DS) \cite{Kritzinger:Werner,Jones:David,Fuller:Aidan}(see Figure \ref{img:dt-types1}). 

A digital model is a static depiction of a physical system or equipment used in design and simulation. Notice that the digital model could exist before the physical entity, e.g., in the design phases of a system. In digital models, there is no automatic data exchange. Therefore, changes in the system do not automatically reflect in the digital world and vice versa. Conversely, a digital shadow is an updated version of a digital model that depicts the physical device or system's current state. In this case, a one-way automatic data flow from the physical to the digital counterpart exists but not the opposite. Finally, a digital twin is a dynamic, real-time representation of a physical device or system that incorporates data from sensors and other sources to deliver an accurate and up-to-date representation. Digital twins can also transmit data to the physical device, providing real-time feedback and control. This level of interactivity makes digital twins a valuable tool even for controlling and optimizations.

\subsection{Technical components}
As previously introduced, a digital twin consists of a physical entity, a virtual representation, and a communication channel between the two. From a technical implementation perspective, the physical entity is generally untouched by the digital twin design. DTs are usually built at the service of the physical entities. Nevertheless, the DT should collect data from the real world to feed the virtual representation. This result in an HRC scenario could be achieved by sensors distributed in the environment, installed on the robot, or worn by humans \cite{huang2021survey}. Depending on the sensors used and the kind of information to extract, the data should go through different processing pipelines before being fed to the virtual environment \cite{qi2021enabling}. Finally, the virtual representation could only consist of mathematical models, but most of the time, programs with a graphical user interface are used for intuitive visualization. The virtual domain of a DT does not only host a mathematical and visual representation of the physical entity. Indeed, it could integrate additional software modules for inferring additional information or planning system behaviors. Therefore, the DT could inject information from the virtual representation into the physical entity, for example, through an augmented reality (AR) interface for humans or direct commands for the robot. To make this possible, the DT shares information over multiple devices, possibly distributed over different networks, as in the case of cloud-based solutions. Therefore, the DT designer should adopt a communication protocol appropriate to the selected application and consider how the overall infrastructure could affect the DT functionalities \cite{lu2020digital}. For example, a DT application using an AR headset will be limited by its computation power, while a DT running the virtual representation on the cloud could have latency issues. Overall, a clear definition and consideration of the technical requirements and constraints regarding a DT implementation are necessary to understand its functionalities and limitations.

\begin{figure*}[t]
\centering
\includegraphics[width=0.7\textwidth]{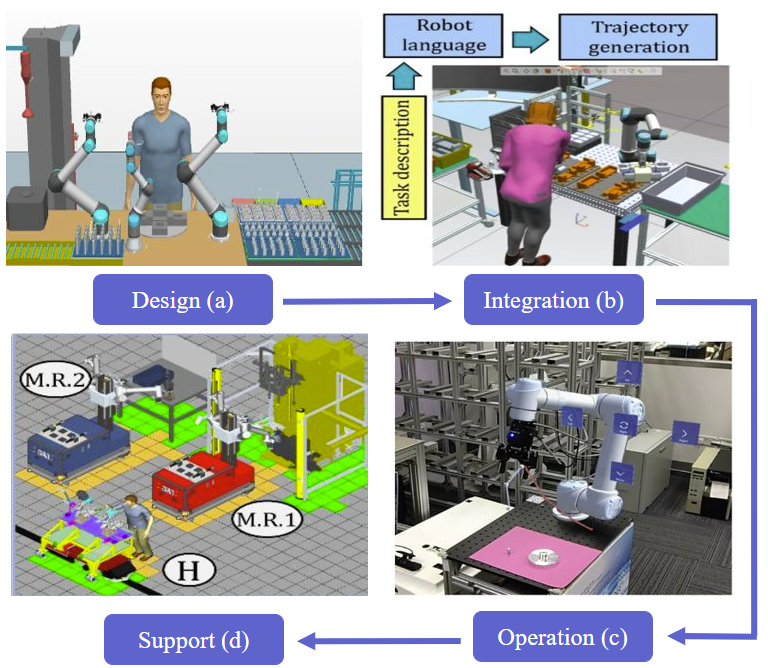}
\caption{A complete life-cycle of a digital twin in HRC. (a) \textit{design phase}: DT for workspace layout design based on vision and object reachability analysis \cite{Malik:Bilberg:dt:2018}, (b) \textit{integration phase}: DT for robot programming and instructions generation  \cite{Malik::Brem}, (c) \textit{operational phase}: DT for robot teleoperation and control through augmented reality \cite{Wang::Xi::Civil}, and \textit{support phase}: (d) DT for workspace reconfiguration \cite{Kousi::Niki}.}
\label{img:dt-types}
\end{figure*}

\section{Digital Twin Taxonomy}
\label{taxonomy}
Based on the definitions collected from the literature and discussed in the previous section, we have designed a taxonomy that would help to better describe the state of the art of digital twins in human-robot scenarios. The taxonomy includes the level of data integration and a new concept, the product life cycle, applied here for the first time to digital twins.

\subsection{Product Life Cycle}

Previous studies have highlighted how digital twins can be introduced and leveraged at different steps of a product life cycle \cite{Jones:David}. We performed a similar analysis for the human-robot collaboration process. Inspired by previous literature \cite{Stark2015}, we defined four life cycles for DT in HRC as design, integration, operation, and support (see Figure \ref{img:dt-types}). As we will see, the distinction between these phases directly influences the role and functionality of a virtual representation. 
 
The design phase is the initial stage of the digital twin lifecycle. In this phase, a virtual replica of a hypothetical physical system is built and validated in terms of design, behavior, layout, and other factors. This step entails creating scenarios for quicker, safer, and better designs of the physical twin, which will be realized later. In this step, the criteria for an effective production system are established and validated, including selecting appropriate tools and robots. Since a physical system does not exist in the design phase, the virtual replica is part of a DM.
 
In the integration stage, the outcomes of the design phase are used to create hardware and software for the physical entity transitioning from digital models to digital shadows or proper digital twins. By incorporating simulation technology in human-robot collaboration, designers can test and optimize the system's performance in a controlled virtual environment, improving efficiency and safety, thus enabling more effective and efficient interaction. This approach provides a cost-effective and efficient means of programming robots for various use cases, thereby enabling highly customized and optimized solutions.

In the operation phase, the collaborative system is in use, and the simulated environment operates in real-time to monitor and regulate the physical twin. Digital twins or shadows are typically used in this phase. The operation phase usually entails gathering and analyzing data from sensors and other sources to monitor the physical system. Therefore, it is the most demanding phase in terms of processing power and network bandwidth. In this phase, the digital twin can be used to discover areas for improvement and make online adjustments to enhance performance or provide visual feedback based on internal information.

Finally, in the support phase, a virtual representation is utilized to support and improve the processes in the physical system. An example is the use of DT to effectively train human operators in a simulated environment or plan workplace adjustments without affecting the physical system.

\section{Literature Review}
\label{review}
We conducted our article search on Scopus using the following query:
\begin{footnotesize}
\begin{verbatim}
    TITLE-ABS-KEY(("digital twin" OR "DT" OR 
    "human W/2 inloop" OR "cyber-physical") AND ("human" OR "operator") 
    AND ("collaboration" OR "collaborative") 
    AND ("assembly" OR "manufacture")) 
\end{verbatim}
\end{footnotesize}
We limited our scope to papers written in English between 2015 and 2022 within the Engineering and Computer Science domain. The search yielded a total of 125 papers, each of which we carefully assessed based on eligibility criteria. We excluded papers that did not involve human-robot collaboration or where DT was not the primary focus, resulting in the exclusion of articles not directly relevant to our study. After the assessment process, only 15 articles remained. The final list of these articles is reported in Table \ref{time}.
%\textcolor{red}{Utilizing Scopus database as our primary information source, we formulated a search query incorporating key terms related to digital twin technology, human-robot collaboration, assembly, and manufacturing. The search query used was: '\textit{ TITLE-ABS-KEY (("digital twin“ OR {DT} OR (human W/2 inloop) OR {cyber-physical}) AND ("human“ OR “operator") AND ("collaboration" OR "collaborative") AND ("assembly" OR "manufacture")) AND (LIMIT-TO (LANGUAGE, "English")) AND (LIMIT-TO (SUBJAREA, "ENGI") OR LIMIT-TO (SUBJAREA, "COMP"))}'. We have limited our scope to English-language papers published between 2015 and 2022. The search yielded a total of 216 papers. We carefully assessed each paper based on  eligibility criteria. We excluded papers that did not involve human-robot collaboration or with DT not being their primary focus, resulting in the exclusion of articles not directly relevant to our research question. We considered various publication types, including articles, conference papers, and reviews. The result of our search is depicted in Table \ref{time}.}

\begin{table}[t]
\caption{Reviewed papers categorized by the level of data integration, considered life cycle, adopted simulator, communication protocol, and architecture nature.\label{time}}
\resizebox{\linewidth}{!}{\begin{minipage}{1.15\linewidth}
\centering
\begin{tabular}{lcccccccc}
\hline
\multirow{2}{*}{Paper} & Data & Life & \multirow{2}{*}{Simulator}      & Communication & \multirow{2}{*}{Architecture}  \\ 
 & Integration & Cycle & & Protocol &\\
\hline
\rowcolor[HTML]{FFFFFF} 
Chancharoen et al., 2022 \cite{Chancharoen::Ratchatin}  & DT  & O  & CoppeliaSim  & Serial & Edge                   \\
\rowcolor[HTML]{FFFFFF} 
Li et al., 2022 \cite{Li::Chengxi::AR} & DS  & I/O & Unity & SSL/TCP  & Cloud                  \\
\rowcolor[HTML]{FFFFFF} 
Li et al., 2022 \cite{Li::Hao::safety} & DT  & O   & Unity & UDP      & Edge               \\
\rowcolor[HTML]{FFFFFF} 
Liu et al., 2022 \cite{Zhang::sun::liu} & DT  & O & -   & OPC-UA  & Edge                 \\
\rowcolor[HTML]{FFFFFF} 
Liu et al., 2022 \cite{Liu::Xinyu::Human-centric} & DS & O & Unity  & TCP  & Cloud               \\   
Gallala et al., 2021 \cite{Gallala::Abir::2022}            & DS     & I        & Unity                                                                                & WebSockets                                                       & Edge                \\
\rowcolor[HTML]{FFFFFF} 
Douthwaite et al., 2021 \cite{Douthwaite::James}    & DT       & D/O   & Unity                                                                      & MQTT                                                             & Edge             \\
\rowcolor[HTML]{FFFFFF} 
Fukushima et al., 2021  \cite{Fukushima::Yuto::Digimobot}      & DS     & I/O    & Unity                                                          & MQTT/WebSocket                                                   & Cloud                 \\
\rowcolor[HTML]{FFFFFF} 
Kousi et al., 2021  \cite{Kousi::Niki}          & DS      & D/O/S  & -                                                           & -                                                      & Edge               \\
\rowcolor[HTML]{FFFFFF} 
Lv et al., 2021 \cite{Qibing::Lv::2021}             & DT   & O     & Tecnomatix                                                                 & TCP                                                          & Edge                 \\
\rowcolor[HTML]{FFFFFF} 
Malik and Brem, 2021 \cite{Malik::Brem}     & DT       & D/I/O & Tecnomatix                                                     & TCP                                                           & Cloud               \\
\rowcolor[HTML]{FFFFFF} 
Tuli et al., 2021 \cite{Tuli::Kohl::Chala}          & DS       & O     & MOSIM                                                               & -                                                      & Edge              \\
\rowcolor[HTML]{FFFFFF} 
Wang et al., 2021 \cite{Wang::Xi::Civil}          & DS      & O     & Unity                                                                           & MQTT                                                             & Cloud               \\
\rowcolor[HTML]{FFFFFF} 
Bilberg and Malik, 2019 \cite{Bilberg:A:DT}    & DT     & O       & Tecnomatix                                                            & -                                                      & Edge               \\
\rowcolor[HTML]{FFFFFF} 
Malik and Bilberg, 2018 \cite{Malik:Bilberg:dt:2018}     & DS   & D/I     & Tecnomatix                          & -                                                     & Cloud                 \\ \hline
\end{tabular}
\end{minipage}}
\end{table}

\subsection{Analysis}

%\textcolor{red}{Considering the definitions provided in section \ref{def} and the presented DT taxonomy, we analyzed the current state-of-the-art on DT for human-robot collaboration.} 
We analyzed the results of our literature search using the presented DT taxonomy as a framework. This analysis aimed to determine how DTs have been adopted and their potential in the HRC. The literature in this regard is young and only a few articles have been published. Furthermore, most studies use the term digital twin regardless of the level of data integration when describing DM or DS. For these reasons, we decided to include in this review articles independently by the level of data integration they consider. In Table \ref{time}, we present a synthesis of the analyses of the 15 papers considered. In the table, for each article, we report the level of data integration adopted and the product life cycle stage in which the simulated environment is used (i.e., D - Design, I - Integration, O - Operation, and S - Support). As we can see in Table \ref{time}, despite all articles using the term digital twin, only seven out of the 15 articles implemented a proper digital twin. We should point out that when multiple virtual representations with different levels of data integration were used, only the highest is reported. %Furthermore, in the table, we also report technical aspects that will be addressed in the discussion (i.e., simulator, communication protocol, and host).

Most of the latter developments in DT for HRC have been achieved in manufacturing contexts \cite{Douthwaite::James,Li::Chengxi::AR}, focusing on assembly scenarios \cite{Malik::Brem,Kousi::Niki,Zhang::sun::liu}, but have also found applications in construction work \cite{Wang::Xi::Civil}. Therefore, most research has focused on robotic manipulators, with little attention paid to mobile robots \cite{Li::Hao::safety}. According to the life cycle in which the simulation environment is integrated, the problems they address and their deployment can vary.
 
In the design phase, digital models are adopted to configure the workspace \cite{Malik:Bilberg:dt:2018} to reduce collisions between humans and robots \cite{Douthwaite::James} while minimizing occlusions and task execution time \cite{Kousi::Niki}. Additionally, human ergonomics and task allocation between agents can be considered at this stage \cite{Malik::Brem}. Following the completion of the design process, the HRC application is alternately integrated into the real world. In this phase, the virtual environment is of significant help in programming the actions and motions of the robots \cite{Malik:Bilberg:dt:2018}. To this extent, different solutions have been proposed leveraging reinforcement learning \cite{Li::Chengxi::AR} and programming by demonstration \cite{Gallala::Abir::2022}. Furthermore, the adoption of DM, DS, and DT allows the simulation of human behaviors, which is helpful for programming more context-aware robot behaviors \cite{Fukushima::Yuto::Digimobot}.
 
Despite the broad use of DM, DS, and DT in all the stages, most research has focused on the operation level. This can be easily observed in Table \ref{time} where 13 out of the 15 articles are associated with the operation phase. In this phase, the virtual representation complements the physical entity with information and models helpful for inferring new details or performing accurate motion and task planning for the robot. Simulation technologies have been leveraged to recognize and predict collisions \cite{Kousi::Niki}, human actions, and human intent \cite{Zhang::sun::liu,Qibing::Lv::2021,Tuli::Kohl::Chala}. Furthermore, for more natural interactions, researchers also explored the monitoring of human ergonomics \cite{Bilberg:A:DT,Malik::Brem,Liu::Xinyu::Human-centric}. This additional information is typically used to influence the PS behavior, e.g., with motion planners reacting to human position in space and planners allocating tasks to the robot according to human behavior \cite{Qibing::Lv::2021,Chancharoen::Ratchatin}. Additionally, virtual representations can send information to the user through communication media such as augmented, mixed, and virtual reality \cite{Wang::Xi::Civil,Douthwaite::James,Li::Hao::safety}. These tools are particularly effective for conveying the human system's intentions or instructions to follow, and are of fundamental importance in teleoperation scenarios \cite{Li::Chengxi::AR}. Finally, limited research attention has been paid to the support phase, in which DS has been adopted to reconfigure the PS \cite{Kousi::Niki}.

In Table \ref{time}, we also report technical details related to the different DM, DS, and DT implementations. In particular, we focused on simulators, communication protocols, and architectures. Regarding the simulators, the most used ones are Unity\footnote{\url{https://unity.com/}} and Tecnomatix\footnote{\url{https://plm.sw.siemens.com/en-US/tecnomatix/}}. However, we should point out that three of the four articles using Tecnomatrix are from the same authors. Therefore the more broadly adopted simulator in the literature is by far Unity. Instead, regarding communication, the most used protocol in the literature were TCP and MQTT. Finally, the last field, architecture, describes if the system has some link with a cloud infrastructure. At this stage, most works in the literature do not integrate their systems into cloud architecture, and those doing that do not take full advantage of the cloud, as we will describe in the next section.

\begin{figure*}[t]
\centering 
\includegraphics[width=0.8 \textwidth]{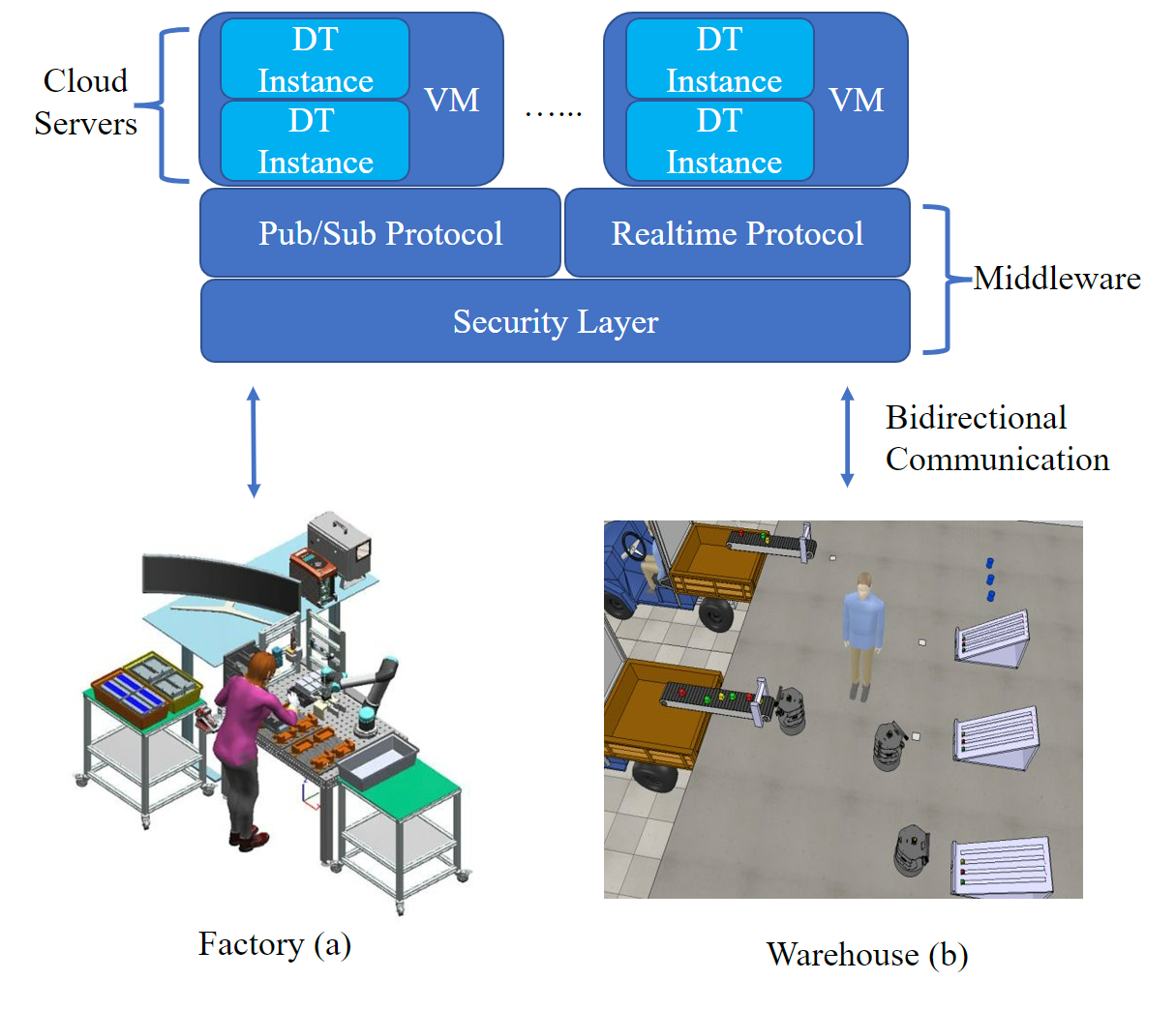}
\caption{A full realization of a cloud-based digital twin for HRC. (a) DT for a factory \cite{Malik::Brem}, and (b) DT for a warehouse \cite{inam2018risk}.}
\label{img:dt-full-real}
\end{figure*}

\section{Challenges and Opportunities}
Digital models are a critical component in the design phase of human-robot collaboration. They allow for a better workspace design through object placement for collision-free cooperation and offline optimization through ergonomic analysis. Serving as a testbed, DMs unlock process optimization through offline task simulation. During the integration phase, there is an evolution toward digital shadows that serve as a virtual environment for human-in-loop robot programming, offline trajectory planning, and workload balancing.
 
The HRC operation phase is often characterized by full-fledged digital twins. In this phase, DT can improve collaboration safety through collision detection and online trajectory planning. The digital twin enables real-time robot teleoperation through various hardware, including AR, VR, and MR, and enables online dynamic workload balancing based on human ergonomics and progress deviation. Furthermore, it can make the robot context and human-aware.
 
In the support phase, DT can provide insights and data analysis, enabling predictive maintenance and task reallocation. They might help redesign the collaborative environment, making it a crucial tool for maximizing efficiency and safety in HRC. Furthermore, novice human operators could use DT for training.
 
In HRC, DT should not be comprised of a single robot-operator couple, but may be extended to multi-agent scenarios in which data must be collected from multiple sources and processed in real-time. In the state-of-the-art, digital twins are often managed by an edge architecture running on hardware with limited computational capabilities, such as head-mounted displays or personal computers. In a realistic scenario, following this paradigm, multiple agent collaboration will encounter processing bottlenecks. On the one hand, the on-premises distribution of the software does not scale horizontally with the load, and on the other hand, utilizing low-processing power devices such as head-mounted displays limits the capability of the digital twin \cite{Li::Chengxi::AR}. For instance, if we consider the dynamic balancing of workload in a factory, the big data managed by the DT is beyond the capability of a single server. Cloud computing has emerged as a potential problem solver that enables the processing and storage of large amounts of data in a horizontally scalable environment; however, its full potential has yet to be realized. Almost half of the papers analyzed utilized the cloud, although still approaching it as a single server without considering scalability, thus not truly benefiting from cloud advantages. 
 
Multiple challenges arise due to the potential of clouds. On top of it are latency and bandwidth. In applications such as teleoperation, safety, and human intention recognition, high communication latency might put operator safety at risk and render DT unusable. Therefore, the choice of communication protocol is critical. MQTT is a low-bandwidth publish/subscribe protocol with high latency \cite{Wang::Xi::Civil}, whereas TCP does not provide sufficiently low latency because of its sensitivity to the network state. Currently, a communication protocol that satisfies latency and bandwidth requirements is yet to be identified for DT. We believe that the third generation of the hypertext transfer protocol (HTTP) is a promising candidate because it offers UDP low latency with TCP packet loss correction and resilience \cite{Saif::Darius}.
 
The security of DT in HRC is yet to be investigated. In HRC, breaching security can result in material damage and human injuries. Gaining access to a fully connected digital twin may affect the behavior of the robot, which may be harmful. Simultaneously, a data breach might expose sensitive information about the operators as well as the product and production process, which are industrial espionage, thus putting them at risk.
 
The realization of a cloud-based digital twin for HRC scenarios may consist of three layers, as depicted in Figure \ref{img:dt-full-real}. In this structure, the middleware consists of a security management layer and a communication protocol, providing publish/subscribe and stream capabilities and allowing event-driven systems and real-time communication. Such a digital twin links all process flows, including factories, assembly lines, and warehouses. This digital twin solution would make optimal use of resources and contribute to cheap ownership costs owing to the horizontal scalability of the cloud infrastructure. With these capabilities, enterprises can improve HRC by increasing efficiency, safety, and productivity in the local workspace and throughout the entire production pipeline.

\section{Conclusion}

A digital twin is a virtual replica of a physical system characterized by bidirectional communication with the physical twin. The importance of digital twins in HRC lies in their ability to offer a comprehensive and accurate representation of real-world systems. This feature enables applications in HRC research fields, such as safety, teleoperation, and human intention recognition. Despite advancements in this domain, DTs suffer scalability issues and are limited by low-performance hardware, such as head-mounted displays.
 
The integration of cutting-edge networking and cloud technologies can further enhance the potential of DTs in HRC. Cloud computing infrastructure offers resources to store, process, and analyze vast amounts of data. Furthermore, cloud technology facilitates collaboration and data sharing among different components of HRC systems, including DTs, robots, and human operators. Cloud computing could also reveal fundamentals for an HRC-enabled production line, from production and assembly to warehouses. This results in increased flexibility and cost-effectiveness \cite{Hu::Liwen::cloud}, but at the same time introduces a set of challenges to be tackled, such as the security and latency of communication.
 
DTs can revolutionize the interaction between humans and robots, leading to new and innovative solutions in various industries and applications. As the demand for HRC increases and systems become increasingly complex, the continued development of DTs and advancements in networking and cloud technology will be crucial to unlocking new possibilities for human-robot collaboration, thus pushing the boundaries of what is possible in the realm of HRC.

\section{Acknowledgment}
This work has been partially supported by the Italian Ministry of University and Research (MUR) through the RAISE project under grant agreement No. D33C22000970006 and by the NVIDIA Academic Hardware Grant Program.
\bibliography{bibliography}

\begin{thebibliography}{10}

\bibitem{ourtzis:d}
D.~Mourtzis, ``Simulation in the design and operation of manufacturing systems:
  state of the art and new trends,'' {\em International Journal of Production
  Research}, vol.~58, no.~7, pp.~1927--1949, 2020.

\bibitem{Chryssolouris:g:ms}
G.~Chryssolouris, {\em Manufacturing systems: theory and practice}.
\newblock Springer Science \& Business Media, 2013.

\bibitem{maccio2022mixed}
S.~Macci{\`o}, A.~Carf{\`\i}, and F.~Mastrogiovanni, ``Mixed reality as
  communication medium for human-robot collaboration,'' in {\em 2022
  International Conference on Robotics and Automation (ICRA)}, pp.~2796--2802,
  IEEE, 2022.

\bibitem{darvish2020hierarchical}
K.~Darvish, E.~Simetti, F.~Mastrogiovanni, and G.~Casalino, ``A hierarchical
  architecture for human--robot cooperation processes,'' {\em IEEE Transactions
  on Robotics}, vol.~37, no.~2, pp.~567--586, 2020.

\bibitem{Kousi::Niki}
N.~Kousi, C.~Gkournelos, S.~Aivaliotis, K.~Lotsaris, A.~C. Bavelos, P.~Baris,
  G.~Michalos, and S.~Makris, ``Digital twin for designing and reconfiguring
  human--robot collaborative assembly lines,'' {\em Applied Sciences}, vol.~11,
  no.~10, p.~4620, 2021.

\bibitem{Malik::Brem}
A.~A. Malik and A.~Brem, ``Digital twins for collaborative robots: A case study
  in human-robot interaction,'' {\em Robotics and Computer-Integrated
  Manufacturing}, vol.~68, p.~102092, 2021.

\bibitem{Bilberg:A:DT}
A.~Bilberg and A.~A. Malik, ``Digital twin driven human--robot collaborative
  assembly,'' {\em CIRP annals}, vol.~68, no.~1, pp.~499--502, 2019.

\bibitem{Kritzinger:Werner}
W.~Kritzinger, M.~Karner, G.~Traar, J.~Henjes, and W.~Sihn, ``Digital twin in
  manufacturing: A categorical literature review and classification,'' {\em
  Ifac-PapersOnline}, vol.~51, no.~11, pp.~1016--1022, 2018.

\bibitem{Grieves::DT::M}
M.~Grieves, ``Digital twin: manufacturing excellence through virtual factory
  replication,'' {\em White paper}, vol.~1, no.~2014, pp.~1--7, 2014.

\bibitem{merlo2023ergonomic}
E.~Merlo, E.~Lamon, F.~Fusaro, M.~Lorenzini, A.~Carf{\`\i}, F.~Mastrogiovanni,
  and A.~Ajoudani, ``An ergonomic role allocation framework for dynamic
  human--robot collaborative tasks,'' {\em Journal of Manufacturing Systems},
  vol.~67, pp.~111--121, 2023.

\bibitem{carfi2021gesture}
A.~Carf{\`\i} and F.~Mastrogiovanni, ``Gesture-based human-machine interaction:
  Taxonomy, problem definition, and analysis,'' {\em IEEE Transactions on
  Cybernetics}, 2021.

\bibitem{Jones:David}
D.~Jones, C.~Snider, A.~Nassehi, J.~Yon, and B.~Hicks, ``Characterising the
  digital twin: A systematic literature review,'' {\em CIRP Journal of
  Manufacturing Science and Technology}, vol.~29, pp.~36--52, 2020.

\bibitem{Fuller:Aidan}
A.~Fuller, Z.~Fan, C.~Day, and C.~Barlow, ``Digital twin: Enabling
  technologies, challenges and open research,'' {\em IEEE access}, vol.~8,
  pp.~108952--108971, 2020.

\bibitem{huang2021survey}
Z.~Huang, Y.~Shen, J.~Li, M.~Fey, and C.~Brecher, ``A survey on ai-driven
  digital twins in industry 4.0: Smart manufacturing and advanced robotics,''
  {\em Sensors}, vol.~21, no.~19, p.~6340, 2021.

\bibitem{qi2021enabling}
Q.~Qi, F.~Tao, T.~Hu, N.~Anwer, A.~Liu, Y.~Wei, L.~Wang, and A.~Nee, ``Enabling
  technologies and tools for digital twin,'' {\em Journal of Manufacturing
  Systems}, vol.~58, pp.~3--21, 2021.

\bibitem{lu2020digital}
Y.~Lu, C.~Liu, I.~Kevin, K.~Wang, H.~Huang, and X.~Xu, ``Digital twin-driven
  smart manufacturing: Connotation, reference model, applications and research
  issues,'' {\em Robotics and Computer-Integrated Manufacturing}, vol.~61,
  p.~101837, 2020.

\bibitem{Malik:Bilberg:dt:2018}
A.~A. Malik and A.~Bilberg, ``Digital twins of human robot collaboration in a
  production setting,'' {\em Procedia manufacturing}, vol.~17, pp.~278--285,
  2018.

\bibitem{Wang::Xi::Civil}
X.~Wang, C.-J. Liang, C.~C. Menassa, and V.~R. Kamat, ``Interactive and
  immersive process-level digital twin for collaborative human--robot
  construction work,'' {\em Journal of Computing in Civil Engineering},
  vol.~35, no.~6, p.~04021023, 2021.

\bibitem{Stark2015}
J.~Stark, ``Product lifecycle management (plm),'' in {\em Product Lifecycle
  Management (Volume 1) 21st Century Paradigm for Product Realisation},
  pp.~1--32, Springer, 2022.

\bibitem{Chancharoen::Ratchatin}
R.~Chancharoen, K.~Chaiprabha, L.~Wuttisittikulkij, W.~Asdornwised, M.~Saadi,
  and G.~Phanomchoeng, ``Digital twin for a collaborative painting robot,''
  {\em Sensors}, vol.~23, no.~1, p.~17, 2022.

\bibitem{Li::Chengxi::AR}
C.~Li, P.~Zheng, S.~Li, Y.~Pang, and C.~K. Lee, ``Ar-assisted digital
  twin-enabled robot collaborative manufacturing system with
  human-in-the-loop,'' {\em Robotics and Computer-Integrated Manufacturing},
  vol.~76, p.~102321, 2022.

\bibitem{Li::Hao::safety}
H.~Li, W.~Ma, H.~Wang, G.~Liu, X.~Wen, Y.~Zhang, M.~Yang, G.~Luo, G.~Xie, and
  C.~Sun, ``A framework and method for human-robot cooperative safe control
  based on digital twin,'' {\em Advanced Engineering Informatics}, vol.~53,
  p.~101701, 2022.

\bibitem{Zhang::sun::liu}
X.~Sun, R.~Zhang, S.~Liu, Q.~Lv, J.~Bao, and J.~Li, ``A digital twin-driven
  human--robot collaborative assembly-commissioning method for complex
  products,'' {\em The International Journal of Advanced Manufacturing
  Technology}, pp.~1--14, 2021.

\bibitem{Liu::Xinyu::Human-centric}
X.~Liu, L.~Zheng, Y.~Wang, W.~Yang, Z.~Jiang, B.~Wang, F.~Tao, and Y.~Li,
  ``Human-centric collaborative assembly system for large-scale space
  deployable mechanism driven by digital twins and wearable ar devices,'' {\em
  Journal of Manufacturing Systems}, vol.~65, pp.~720--742, 2022.

\bibitem{Gallala::Abir::2022}
A.~Gallala, A.~A. Kumar, B.~Hichri, and P.~Plapper, ``Digital twin for
  human--robot interactions by means of industry 4.0 enabling technologies,''
  {\em Sensors}, vol.~22, no.~13, p.~4950, 2022.

\bibitem{Douthwaite::James}
J.~A. Douthwaite, B.~Lesage, M.~Gleirscher, R.~Calinescu, J.~M. Aitken,
  R.~Alexander, and J.~Law, ``A modular digital twinning framework for safety
  assurance of collaborative robotics,'' {\em Frontiers in Robotics and AI},
  vol.~8, p.~758099, 2021.

\bibitem{Fukushima::Yuto::Digimobot}
Y.~Fukushima, Y.~Asai, S.~Aoki, T.~Yonezawa, and N.~Kawaguchi, ``Digimobot:
  Digital twin for human-robot collaboration in indoor environments,'' in {\em
  2021 IEEE Intelligent Vehicles Symposium (IV)}, pp.~55--62, IEEE, 2021.

\bibitem{Qibing::Lv::2021}
Q.~Lv, R.~Zhang, X.~Sun, Y.~Lu, and J.~Bao, ``A digital twin-driven human-robot
  collaborative assembly approach in the wake of covid-19,'' {\em Journal of
  Manufacturing Systems}, vol.~60, pp.~837--851, 2021.

\bibitem{Tuli::Kohl::Chala}
T.~B. Tuli, L.~Kohl, S.~A. Chala, M.~Manns, and F.~Ansari, ``Knowledge-based
  digital twin for predicting interactions in human-robot collaboration,'' in
  {\em 2021 26th IEEE International Conference on Emerging Technologies and
  Factory Automation (ETFA)}, pp.~1--8, IEEE, 2021.

\bibitem{inam2018risk}
R.~Inam, K.~Raizer, A.~Hata, R.~Souza, E.~Forsman, E.~Cao, and S.~Wang, ``Risk
  assessment for human-robot collaboration in an automated warehouse
  scenario,'' in {\em 2018 IEEE 23rd International Conference on Emerging
  Technologies and Factory Automation (ETFA)}, vol.~1, pp.~743--751, IEEE,
  2018.

\bibitem{Saif::Darius}
D.~Saif and A.~Matrawy, ``Apure http/3 alternative to mqtt-over-quic in
  resource-constrained iot,'' in {\em 2021 IEEE Conference on Standards for
  Communications and Networking (CSCN)}, pp.~36--39, IEEE, 2021.

\bibitem{Hu::Liwen::cloud}
L.~Hu, N.-T. Nguyen, W.~Tao, M.~C. Leu, X.~F. Liu, M.~R. Shahriar, and S.~N.
  Al~Sunny, ``Modeling of cloud-based digital twins for smart manufacturing
  with mt connect,'' {\em Procedia manufacturing}, vol.~26, pp.~1193--1203,
  2018.

\end{thebibliography}
\bibliographystyle{ieeetr}

\end{document}